\documentclass[final]{cvpr}

\usepackage{times}
\usepackage{epsfig}
\usepackage{graphicx}
\usepackage{amsmath}
\usepackage{amssymb}
\usepackage{xcolor}
\usepackage{tabularx} 
\usepackage[lofdepth,lotdepth]{subfig}
\usepackage{enumitem}
\usepackage{booktabs}
\usepackage{arydshln}
\usepackage{subfig}
\usepackage[switch]{lineno}

\usepackage[pagebackref=true,breaklinks=true,letterpaper=true,colorlinks,bookmarks=false]{hyperref}

\def\cvprPaperID{****} 

\begin{document}

\title{NUTA: Non-uniform Temporal Aggregation for Action Recognition}

\author{
Xinyu Li\thanks{equally contribute}, Chunhui Liu\footnotemark[1], Bing Shuai, Yi Zhu, Hao Chen, Joseph Tighe\\
Amazon Web Services \\
{\tt\small  \{xxnl, chunhliu, bshuai, yzaws, hxen, tighej\}@amazon.com}

}

\maketitle
\begin{abstract}
In the world of action recognition research, one primary focus has been on \textbf{how to} construct and train networks to model the spatial-temporal volume of an input video. These methods typically uniformly sample a segment of an input clip (along the temporal dimension).  However, not all parts of a video are equally important to determine the action in the clip. In this work, we focus instead on learning \textbf{where to} extract features, so as to focus on the most informative parts of the video. 
We propose a method called the non-uniform temporal aggregation (NUTA), which aggregates features only from informative temporal segments. We also introduce a synchronization method that allows our NUTA features to be temporally aligned with traditional uniformly sampled video features, so that both local and clip-level features can be combined. Our model has achieved state-of-the-art performance on four widely used large-scale action-recognition datasets (Kinetics400, Kinetics700, Something-something V2 and Charades). In addition, we have created a visualization to illustrate how the proposed NUTA method selects only the most relevant parts of a video clip.
\end{abstract}

\section{Introduction}

A key challenge in action recognition is how to learn a feature that captures the relevant spatial and motion queues in an efficient and compact representation. 
This problem has been well studied, primarily using convolution neural networks (CNNs) \cite{he2016deep}, from frame-based methods \cite{karpathy2014large} to segment based temporal information aggregation \cite{wang2016temporal} to the I3D based methods \cite{carreira2017quo}. These methods have primarily focused on \textbf{how to} perform feature extraction that captures a complete spatial-temporal description of the video. The majority of these methods treat each frame, or point in time, with equal weight, but not all parts of the video are equally important and thus it is also key that we develop feature extraction methods that can determine \textbf{where to} extract features from. 
Some recent works have started to look at this problem with non-local modules \cite{wang2018non}, directional convolutions \cite{li2020directional} or ensembles of features at different sample rates \cite{feichtenhofer2018slowfast, yang2020temporal}, but these methods still focus more on how to comprehensively extract features rather than where to extract the features. In this work we leverage all the advancements that have been made in the past years on how to extract motion features and focus our attention on the question of \textbf{where to} extract these features from. 
%

There are two major challenges to answer the ``\textbf{where to}'' question.
First, non-uniformly extracting features efficiently from only informative temporal episodes is challenging as it is required to look at the whole video to determine which parts are informative.
Some recent work \cite{korbar2019scsampler, wu2019multi} have proved that a recognition system can benefit from selecting the informative frames rather than simply taking the uniformly sampled frames as inputs. However, these systems treat the frame selection and feature extraction as two separate stages and thus the frame selection can not benefit from the later feature extraction thus reducing the descriptive power of the network and adding redundancy in the two stages.
\begin{figure}[t]
	\begin{center}
		\includegraphics[width=1\columnwidth]{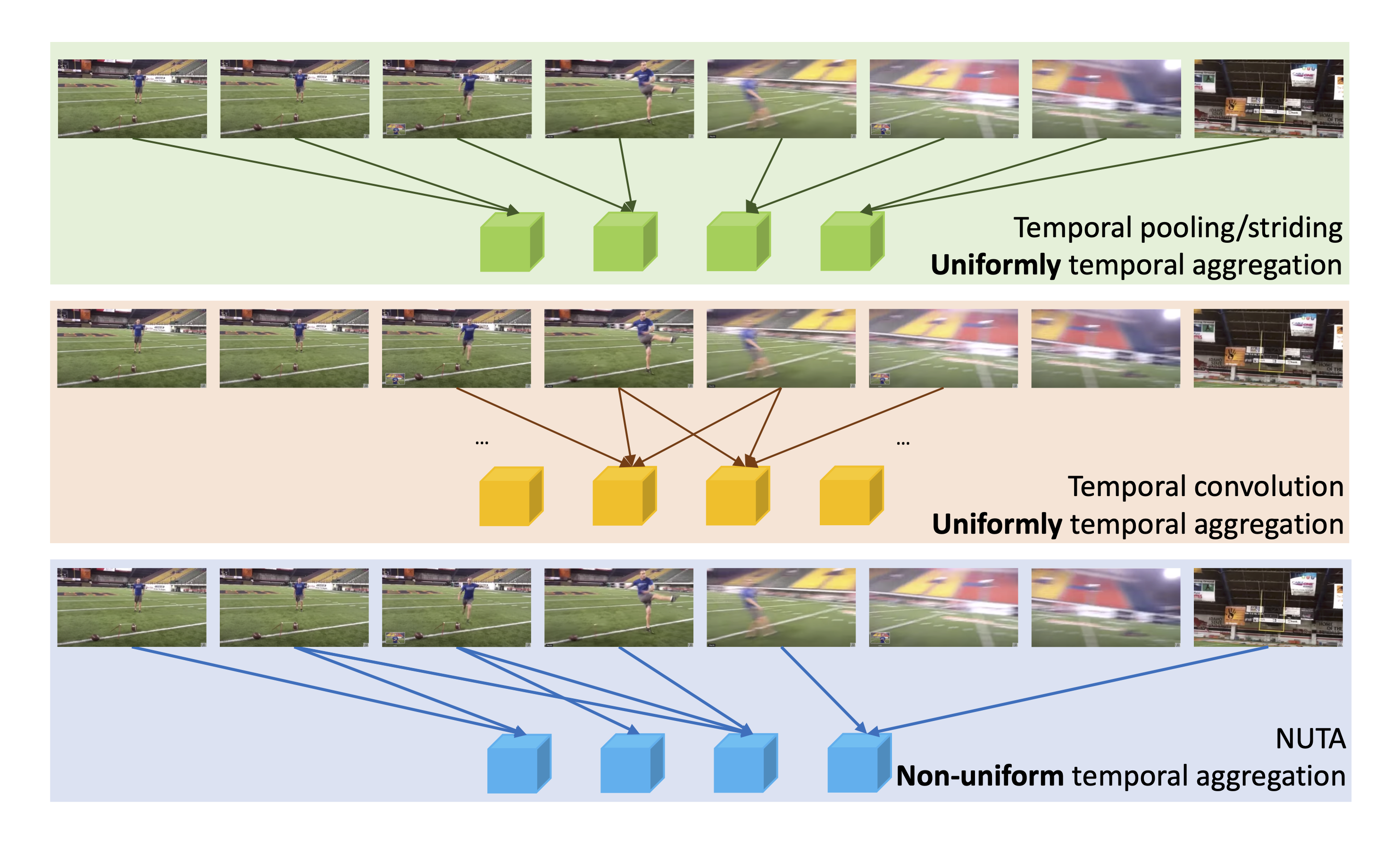}
	\end{center}
	\caption{
	A comparison of different temporal modeling methods. The temporal pooling/striding and temporal convolution aggregate features uniformly along time. Our non-uniform temporal aggregation (NUTA) extracts features non-uniformly from informative parts of the video clip.
	}
	\label{fig:1}
\end{figure}
Second, exchanging information between different feature sampling strategies efficiently and comprehensively remains a challenge.
Other works have tackled this using kernels \cite{li2020tea, liu2020tam, wang2018non}, LSTMs \cite{du2017recurrent} or most recently feature pyramids \cite{yang2020temporal} but each of these methods only partially deals with this information exchange problem. 

In order to address the first challenge (how to extract feature from only representative temporal instances), 
we introduce a non-uniform temporal aggregation (NUTA) method. In contrast to the average pooling operator or temporal stride convolution kernels (Figure \ref{fig:1}), the NUTA module generates a temporal map that non-uniformly projects $T$ temporal instances to $T'$ ($T' < T)$.
The NUTA has receptive field in temporal dimension as large as the entire clip so the NUTA module is able to generate clip-level features.
Differing from previous two-stage systems that perform frame selection first and then do recognition on the selected frames \cite{korbar2019scsampler, wu2019multi}, our NUTA module performs temporal information selection at the feature level in an end-to-end manner. 
Differing from previous work that focus on ``stacking'' features of the same kind \cite{feichtenhofer2018slowfast,wang2018non}  to construct the comprehensive feature representation, NUTA works by focusing on a sub-set of the frames that are the most informative for the task.
To address the second challenge (information exchange between network branches), we propose a two-branch network with a novel temporal synchronization strategy. 
The two-branch design maintains the uniformly aggregated local feature branch and the non-uniformly aggregated clip-level feature branch separately. 
Note that the purpose of our two-branch design is to separate the feature descriptors instead of using multiple input modalities to encourage different feature extraction behaviors as in \cite{feichtenhofer2018slowfast,yang2020temporal}.
The temporal synchronization strategy temporally aligns the features from the two branches to allow the information from each branch to flow to the other.
Such design not only improves the action recognition performance by removing noise from both the uniform and non-uniform branches, but also reduces the computational requirements of our network, allowing our two-branch design to maintain similar computational cost to the original single branch network.

We tested our model on four of the most popular action recognition datasets: Kinetics 400, Kinetics 700, Something Something V2 and Charades.
Our NUTA network achieves the state-of-the-art (SOTA) or comparable performance on all four datasets with less computation cost. 
We perform a detailed ablation study that shows our non-uniform aggregation method is effective across a wide range of configurations.
We also visualized our model to better demonstrate how our NUTA behaves and how it helps answer the ``where to'' question. To summarize, our contributions are:
\begin{enumerate}[itemsep=0pt,parsep=0pt]
\item We propose a NUTA module to answer the ``where to'' question, which is able to identify and sample only the informative parts of the video clips;
\item We propose a two-branch NUTA network for action recognition, which learns local and clip-level features by fusing the feature extracted from the best in bread video modeling methods and our novel NUTA module; 
\item We propose information fusion and temporal synchronization methods that allow both branches to specialize on their own aspects of video modelling while sharing their feaures after each stage of the network; 
\item We present experimental results on four common, large scale action recognition datasets, which shows state-of-the-art or better performance on each. 
\end{enumerate}

\section{Related Work} 
\label{sec:related_work}
The early video action recognition work used 2D CNNs  \cite{karpathy2014large,simonyan2014two} to extract and aggregate frame-wise features with no effective temporal modeling. 
Therefore, the question of how to perform temporal modeling became a key area of interest.
RNNs and LSTMs \cite{hochreiter1997long} were used to model temporal associations among features extracted by 2D networks \cite{li2017progress,yue2015beyond,donahue2015long,li2018videolstm, hu2017temporal, liu2017online}.
However, LSTM based methods often significantly increased the computation with little benefit to accuracy.
Segment level predictions methods such as TSN \cite{wang2016temporal} and rank pooling \cite{fernando2016rank,bilen2016dynamic} based methods were proposed to better capture the temporal feature evolution at the video level. 
Although quite successful, these methods still followed the idea of aggregating 2D features instead of performing pixel-level 3D feature learning.
More recently, 3D CNNs \cite{carreira2017quo,taylor2010convolutional,tran2018closer,xie1712rethinking,tran2015learning, qiu2017learning,martinez2019action,tran2018closer} have gained significant popularity due to their ability to perform spatio-temporal feature modeling. 
The 3D CNNs effectively address the ``how to'' question and are now used as basic building blocks for most temporal modeling tasks.
Most of recent works have continued to focus on ``how to'' perform temporal modeling more comprehensively: the non-local network \cite{wang2018non} leverages non-local connections to establish pixel level long-term spatio-temporal association; the TEA \cite{zhou2018temporal} and TAM \cite{liu2020tam} methods use a local and global kernel for long-shot term feature learning; the SlowFast \cite{feichtenhofer2018slowfast} and TPN \cite{yang2020temporal} methods ensemble feature at different frame-rate for better performance.
There have been a handful of papers that have tried to answer the ``where to'' question.
Early research tried to generate long-term feature with LSTM and temporal attention \cite{song2018spatio,du2017recurrent} but none of these methods ended up achieving competitive performance.
Two recent methods, SCSampler \cite{korbar2019scsampler} and multi-agent sampler \cite{wu2019multi} select only informative clips from a video for action recognition and have been able to demonstrate better performance than uniformly sampling the frames from a video. 

Different from  these previous work, we answer the ``where to'' question by proposing the non-uniform temporal aggregation module which learn the feature non-uniformly from informative temporal episodes. Instead of adding features, e.g. from non-local connection or from multiple branches that takes different types of input \cite{wang2018non,feichtenhofer2018slowfast}, our network reduce redundant information for better recognition. Different from two-stage system that first select representative frames and run classifier on selected frames \cite{korbar2019scsampler}, our proposed method runs end-to-end and performs feature level information selection.

Many two-branch architectures have been proposed to fuse information from different sources, e.g. input frames at different resolution \cite{karpathy2014large}, RGB and optical flow images \cite{wang2016temporal,simonyan2014two,feichtenhofer2017spatiotemporal}, and video clips at different sample rate \cite{feichtenhofer2018slowfast, zhou2018temporal}.
Differing from most previous two-branch systems that fuse features from different input modalities or representations, our two-branch design models both different and complementary features from the same input.

\section{Methodology}
\label{sec:method}
\begin{figure*}[t]
\centering
\includegraphics[width=1\textwidth]{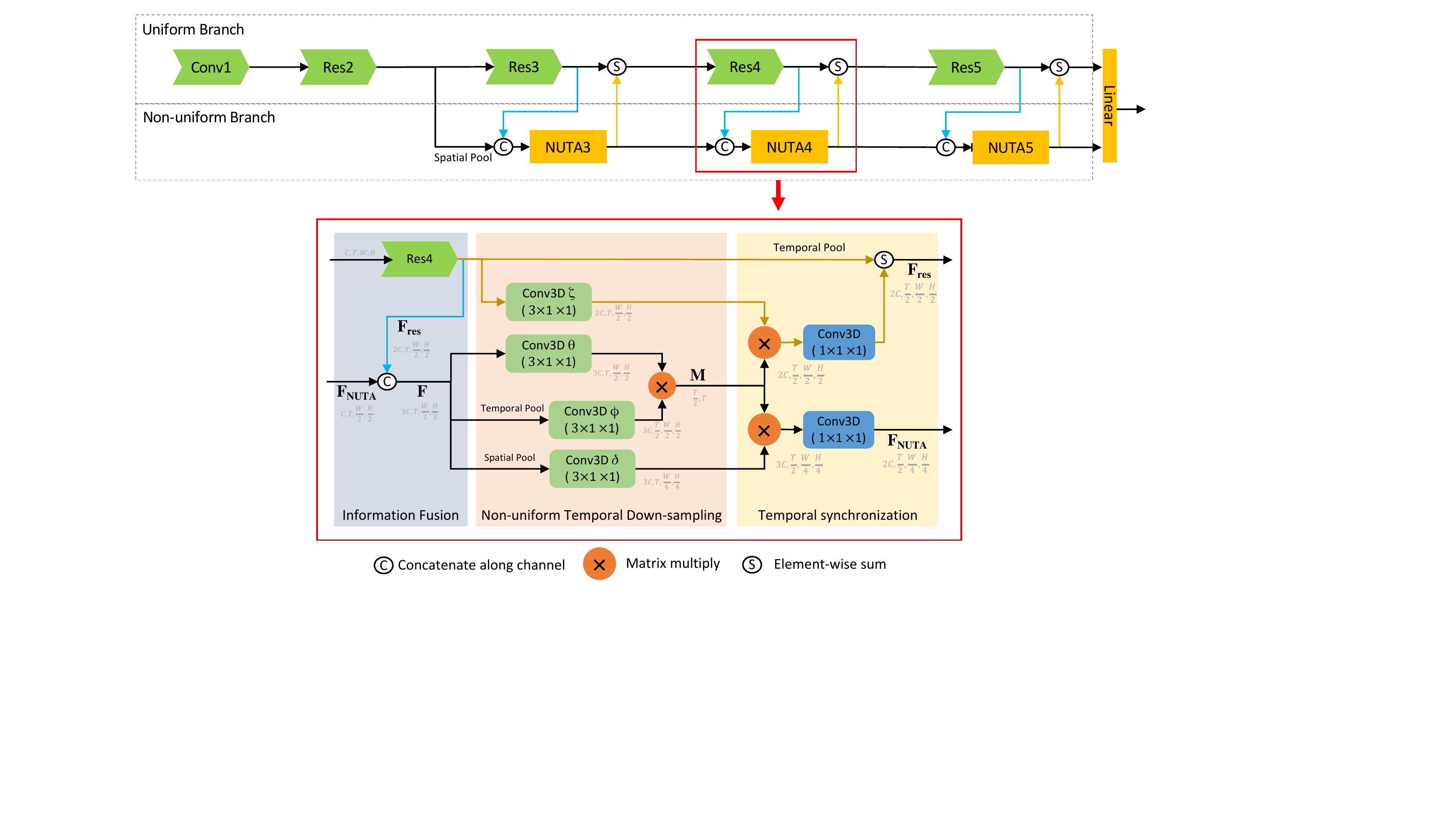}
\caption{An overview of proposed NUTA network: a two-branch design (top). A detailed view of our NUTA module and its interaction with uniform branch is show on the bottom.
}
\label{fig:2}
\end{figure*}

\subsection{Overview}
We propose the Non-uniform Temporal Aggregation (NUTA) network that follows a two-branch design with one uniform branch that maintains the relative temporal spacing of the input clip and captures temporally local information, and one non-uniform branch for clip-level temporal feature selection (Figure \ref{fig:2}). 
We adopt a 3D convolutional style network for our uniform branch to learn local spatial-temporal features. 
We propose a novel NUTA module and build our non-uniform branch by stringing multiple NUTA modules together. 
Each NUTA module is able to model the full clip-level temporal feature by sampling non-uniformly along the temporal dimension. 
Our NUTA module does not perform any spatial modeling, thus to enable the non-uniform branch to model the clip at different spatial scales we transfer information between the two branches after each stage. 
This information exchange between our uniform and non-uniform branches is non trivial because each operates on different samples from the input clip. 
To overcome this issue we introduce our temporal synchronization strategy, which synchronizes the features from the uniform and non-uniform branches bidirectionally. 
The rest of this section introduces the components of our NUTA module including the non-uniform down-sampling, feature synchronization and feature fusion strategies.

\subsection{Non-uniform Temporal Down-sampling}
\label{sec:NUTA}

A core feature of our NUTA network is our non-uniform sampling of the video clip feature along the temporal dimension. This component is to only sample the most informative parts to feature across time. We achieve this via our down-sampling module that is parameterized as a learned projection map, which maps a feature of temporal dimension $T$ to $T'$ ($T'=\frac{T}{2}$ in this paper). Given a feature tensor $\mathbf{F} \in \mathbb{R}^{ C \times T \times W \times H}$, 
we apply a self-attention-like module to learn the projection map $\mathbf{M}$ as:
\begin{equation}
	\mathbf{M}= \operatorname{softmax}\left(\Gamma[\phi(\mathbf{\operatorname{pool}(F}))] \times \Gamma[\theta(\mathbf{F})]^T\right)
\label{eq:temp_conv}
\end{equation}
where $\operatorname{pool}(*)$ denotes temporal max-pooling with kernel size of 2, functions $\phi(*)$ and $\theta(*)$ are convolution operators, and $\Gamma[*]$ denotes a reshape and permutation operation.
Then, the clip-level, non-uniformly sampled feature $\mathbf{F_{NUTA}}$ is generated by applying the temporal projection map as:
\begin{equation}
	\mathbf{F_{NUTA}}=  \operatorname{Conv}\left( \Gamma^{-1}[\mathbf{M} \times \Gamma [\delta(\mathbf{F})]] \right)
\label{eq:temp_attn}
\end{equation}
where the $\delta(*)$ denotes a convolution, the $\Gamma^{-1}$ is the inverse reshape of $\Gamma$, and $\operatorname{Conv}$ denotes a convolution to compress the channel dimension for information fusion with the next stage. In practice, we use a $(3,1,1)$ convolution kernel for $\phi(*)$, $\theta(*)$ and $\delta(*)$. We find such convolution gives a better performance than commonly used $(1, 1, 1)$ kernel, which may due to it gives larger capacity on temporal modeling.

Depending on the complexity of the scene and action classification needed, there may be some set of frames that are most informative for one type of action and a completely different set for another type of action.
To allow our network to have the flexibility of picking different sets of temporal periods for each channel we extend our projection map by making it multi-head. Each head is then free to select its own set of frames and extract features from those frames.
We implement this using our $\Gamma[*]$ operation that split $\phi(\mathbf{F}) \in \mathbb{R}^{T \times C \times W \times H}$ to $n$ heads along the channels, such that $\Gamma[\phi(\mathbf{F})] \in \mathbb{R}^{ n \times  T \times \frac {C}{n}WH}$. 
To make our design computationally efficient, we use the group convolution and empirically set groups as 64 (see section ablation study for details).

Although it may appear similar, our proposed temporal modeling is significantly different from non-local network \cite{wang2018non} as our generated projection map $\mathbf{M} \in \mathbb{R}^{n \times \frac T 2 \times T }$ represents a temporal association based on spatial features while the non-local kernel $\mathbf{M'} \in \mathbb{R}^{TWH \times TWH}$ represents a pixel level spatial temporal association learned from channels. 

\subsection{Temporal Synchronization}
\label{sec:temp_sync}
The output feature from our NUTA module $\mathbf{F_{NUTA}}_{n} \in  \mathbb{R}^{C \times \frac{T}{2} \times W \times H }$ is not compatible with the uniform branch feature ($\mathbf{F_{res}}_{n} \in  \mathbb{R}^{C \times T \times W \times H }$) for two reasons:
1) the temporal dimensions do not match ($\frac{T}{2} \neq T$) and 2) they are not sampled from the same set of frames. 
To handle this incompatibility we propose to synchronize the features from the non-uniform branch to the uniform branch (Figure \ref{fig:2} yellow arrow) as:
\begin{equation}
	\mathbf{\mathbf{F}_{res}}= \operatorname{Conv}\left( \Gamma^{-1}[\mathbf{M} \times \Gamma [\zeta(\mathbf{F_{res}})]] \right)+\operatorname{pool}(\mathbf{F_{res}})
\label{eq:temp_sync}
\end{equation}
where $\zeta$ denotes a 3D convolution with kernel $(3,1,1)$ similar to $\theta$ and $\phi$. 
Operation $\operatorname{Conv}\left( \Gamma^{-1}[\mathbf{M} \times \Gamma [\zeta(\mathbf{F_{res}})]] \right)$ can be intuitively understood as applying the same temporal projection to the uniform branch feature $\mathbf{F_{res}}$, so that feature is synchronized with clip-level feature from the non-uniform branch.
Note that although we apply the same temporal projection map to the feature from uniform branch, the following 3D resnet layers will still generate the local features. 
The temporal synchronization removes the non-informative features from uniform branch, which helps improve the action recognition performance and reduces the computation cost (see ablation study section).

\subsection{Information Fusion}
\label{sec:info_fus}
The temporal synchronization handles the synchronization and combination of the non-uniform features with the uniform branch, however it doesn't encode the spatial information learned in the uniform branch into the non-uniform branch. To address this, we fuse the uniform branch feature from res-stage $n$ and non-uniform feature from NUTA-stage $n-1$ and feed them to NUTA-stage $n$ as follows:
\begin{equation}
	\mathbf{F}_n= f[\mathbf{F_{NUTA}}_{n-1},\mathbf{F_{res}}_{n}]
\label{eq:info_excha}
\end{equation}
where $	\mathbf{F}_n$ stands for the fused feature at $n_{th}$ stage. $\mathbf{F_{NUTA}}_{n-1}$ denotes the feature from non-uniform branch (NUTA module) and $\mathbf{F_{res}}_{n}$ is the feature from the uniform branch (ResNet feature backbone).
We consider the following three fusion strategies $f$:
(1). Non-local fusion ($f_n$): where we use $\mathbf{F_{NUTA}}_{n-1}$ as query to pull information from $\mathbf{F_{res}}_{n}$.
(2). Feature sum-up ($f_s$): where we apply the 1D convolution to  $\mathbf{F_{NUTA}}_{n-1}$ and apply the pixel wise addition to $\mathbf{F_{res}}_{n}$.
(3). Channel concatenate ($f_c$): where we concatenate $\mathbf{F_{res}}_{n}$ with $\mathbf{F_{NUTA}}$ along channel dimension.
Our results show that $f_c$ gives best performance at lowest computational cost.
Note that the initial $\mathbf{F_{NUTA}}$ is generated by spatially down-sampling the uniform branch feature as shows in Figure \ref{fig:2}. 

\subsection{Implementation Details}
\label{sec:implementation}
We initialize the uniform branch with Res2D or I3D backbone networks using ImageNet pre-trained weights.
Our experiments show that the training process converges faster if the uniform branch is initialized with Kinetics 400 pre-trained weights.
For the other components, we used kaiming initialization \cite{he2016deep}.
The model is trained on 8 machines with 8 GPUs on each machine (64 Tesla V100 GPUs in total). 
The batch size is set to 8 per GPU without using the synchronized batch normalization.
If the batch size is smaller than 8 per GPU, we train the model with synchronized batch normalization.

All models are trained for 180 epochs. 
We use an initial learning rate of 0.4 (equivalent to 0.05 on a single instance) and weights decay of 1e-4. 
Learning rate is dropped by scale of 10 at epoch 60, 120 and 150.
To avoid over-fitting, we apply the commonly used data augmentation techniques including: random resized crop (i.e., with short side randomly resized to [256, 320] and random crops of 224 following \cite{simonyan2014very,wang2018non}), random temporal sampling (i.e., we keep the original video frame rate, when a video has less frames than needed, we pad the missing frames with last frame) and random horizontal flipping. 
Besides data augmentation, we also use dropout \cite{hinton2012improving} before the final linear layer with a dropout ratio of $0.6$.

\section{Experimental Results}
\label{sec:exp}

\subsection{Dataset}
\label{sec:dataset}

We test our method on four of the most widely used datasets: \\
\textbf{Kinetics 400 \cite{kay2017kinetics}} consists of approximately 240k training and 20k validation videos trimmed to 10 seconds from 400 human action categories. Similarly to other works, we report top-1 classification accuracy on the validation set.\\
\textbf{Kinetics 700 \cite{carreira2019short}} contains approximately 650k video clips that covers 700 human action classes. Each action class has at least 600 video clips. Each clip is human annotated with a single action class and lasts around 10s. We report top-1 classification accuracy on the validation set.\\
\textbf{Something-Something V2 \cite{goyal2017something}} dataset consists of 174 actions and contains 220,847 videos. Following other works, we report top-1 classification accuracy on the validation set. Something-Something dataset is fairly different from the other three datasets as it does not depict people performing common activities but instead manipulations of one or two objects such as opening something, covering something with something, moving something behind something. This dataset in particular requires strong temporal modeling as many activities cannot be simply inferred based on spatial features. \\
\textbf{Charades \cite{sigurdsson2016hollywood}} has about 9.8k training videos and 1.8k validation videos in 157 classes in a multi-label classification setting with longer activities spanning about 30 seconds on average. Performance is measured in mean Average Precision (mAP).
\begin{table}[t!]
\caption{Results on Kinetics-400 dataset. We report top 1(\%) on the validation set. I3D* stands for the I3D network with temporal down sampling \cite{wang2018non}. The `Input' column indicates what frames of the 64 frame clip are actually sent to the network. $n \times \tau$ input indicates we feed $n$ frames to the network sampled every $\tau$ frames.}
	\begin{center}
		\begin{tabularx}{1\columnwidth}{l|c|c|c} 
		    \toprule
			Model        & Input & GFLOPs & Top1  \\ 
			\midrule
			R2D50 \cite{wang2018non}   & $32\times2 $   & - & 69.9  \\
			R2D101 \cite{wang2018non}   & $32\times2 $   & - & 71.3  \\
			R2D50-NL \cite{wang2018non}   & $32\times2 $   & - & 73.5  \\
			TSM \cite{lin2018temporal}     & 8   & 33 & 74.1  \\
			\midrule
			NUTA R2D-50              & $32\times2$ & 54 & 74.2 \\
			NUTA R2D-101             & $32\times2$ & 66 & 75.5 \\
			\midrule
			\midrule
			I3D50* \cite{carreira2017quo}   & $32\times2$   & 33 & 74.0  \\
			I3D101* \cite{carreira2017quo}  & $32\times2$   & 58 & 77.4  \\
			I3D50 \cite{yang2020temporal}  & $32\times2$   & 168  & 75.7  \\
			NL50 \cite{wang2018non}    & $32\times2$   & 282 & 76.5  \\
			NL101 \cite{wang2018non}   & $32\times2$   & 544 & 77.7  \\
			TEA50 \cite{li2020tea}   & $16\times2$   & 70 & 76.1  \\
			CIDC \cite{li2020directional}   & $32\times2$   & 101 & 75.5  \\
			FG 152 \cite{Martinez_2019_ICCV}   & $32\times2$   & - & 78.8  \\
			SF50 \cite{feichtenhofer2018slowfast}     & $32\times2$   & 66 & 77.0  \\
			SF101 \cite{feichtenhofer2018slowfast}   & $32\times2$   & 106 & 77.5  \\
			TPN-50 \cite{yang2020temporal}   & $32\times2$   & 256 & 77.7  \\
			TPN-101 \cite{yang2020temporal}   & $32\times2$   & 458 & 78.9  \\  
			X3D-L \cite{feichtenhofer2020x3d}   & $16\times5$   & 24.8 & 77.5  \\ 
			\midrule
			NUTA network I3D-50*              & $32\times2$ & 50 & 76.2 \\
			NUTA network I3D-50              & $16\times4$ & 98 & 76.8 \\
			NUTA network I3D-50              & $32\times2$ & 197 & 77.2 \\
			NUTA network I3D-101*             & $32\times2$ & 111 & 77.6 \\
			NUTA network I3D-101             & $16\times4$ & 186 & 78.3 \\
			NUTA network I3D-101              & $32\times2$ & 372 & 78.9 \\
			\bottomrule
		\end{tabularx}
	\end{center}
	\label{tab:k400_res}
\end{table}
\subsection{Comparison to State-of-the-art}
\label{tab:main_res}

\subsubsection{Kinetics 400} 
Following previous works, at inference we run 30-view (10 temporal crops and 3 spatial crops) through our network and average their predictions to procure the final classification score. We report results on the validation set of Kinetics 400 in Table \ref{tab:k400_res}. We report the top 1 accuracy and GFLOPs (Giga Floating-Point Operations) required to compute results on one view.

Our method achieves comparable performance to state-of-the-art methods with fewer FLOPs and notably outperforms other methods that have a similar goal of answering where to focus the feature extraction across the video clip.
In comparison to the non-local model that emphasizes long-term temporal modeling \cite{wang2018non}, our network achieves 1.2\% higher accuracy with roughly half of the computation. Looking at our I3D-50 variant, we outperform the most recent TEA network \cite{li2020tea} by 0.7\% at roughly same FLOPs using 16 frames input. 
Compared to state-of-the-art networks \cite{feichtenhofer2018slowfast, yang2020temporal}, our model achieves similar performance with 30\% less FLOPs compared with TPN \cite{yang2020temporal}. 
The results demonstrate that dropping non-informative frames is an efficient way for temporal feature aggregation in comparison to fusing features at multiple different scales.
The proposed NUTA network boosts the performance of an I3D network by 3.2\% but only increases computation by about 8\%, which demonstrates the proposed network architecture is computationally efficient.

NUTA is generalizable to different backbones. From the table we show that NUTA network works well with both the I3D variant that includes temporal down-sampling \cite{wang2018non} and the one that has no temporal down-sampling \cite{yang2020temporal}. 
Of the methods in Tabel \ref{tab:k400_res}, X3D \cite{feichtenhofer2020x3d}, which is based on neural architecture search, arguably has the best trade-off between performance and cost. Our NUTA network is fully compatible with it and other 3D convolution based backbones as well. 

The NUTA module also works with 2D backbones. 
The results in the top rows of Table \ref{tab:k400_res} show that our model is able to outperform the recently proposed 2D TSM  \cite{lin2018temporal} network with the similar computational cost, which demonstrates that the non-uniform temporal information aggregation is effective on both 2D and 3D feature extraction pipelines.

\subsubsection{Kinetics 700}
Kinetics 700 \cite{carreira2019short} is the latest Kinetics dataset for video classification. 
We follow the same 30-view evaluation protocol as used in Kinetics 400 and report the FLOPs and top1 accuracy.

\begin{table}[t!]
\caption{Results on Kinetics-700 dataset. We report top 1 accuracy (\%) on validation set. K4\&K6 stand for Kinetics 400  and Kinetics 600. SF-NL 101 stands for slowfast network with non-local block \cite{feichtenhofer2018slowfast}. $n \times \tau$ indicates we feed $n$ frames to the network sampled every $\tau$ frames. }
	\small
	\begin{center}
		\begin{tabularx}{\columnwidth}{l|c|c|c} 
		    \toprule
			Model        & pre-train & FLOPs & Top1 \\
			\midrule
			I3D50 \cite{carreira2019short}    & N/A & N/A   & 58.7  \\
			SF101-NL $8\times8$     & K4\&K6   & 115G & 70.6 \\
 			SF101-NL $16\times8$   & K4\&K6   & 234G & 71.0 \\
		    \midrule
			NUTA network I3D-50   ($32\times2$)               & K400  & 197G & 68.9 \\
			NUTA network I3D-101  ($16\times4$)              & K400  & 186G & 69.5 \\
			NUTA network I3D-101  ($32\times2$)            & K400  & 372G & 70.3 \\
			\bottomrule
		\end{tabularx}
	\end{center}
	\label{tab:k700_res}
\end{table}

Our experiments show a consistent performance trend on Kinetics 700 dataset. As listed in Table \ref{tab:k700_res},
our model achieves comparable performance to slowfast \cite{feichtenhofer2018slowfast}, and significantly outperforms the baseline I3D by 10\%.
Note that the slowfast model is pre-trained on both Kinetics 400 and 600 while our model is only pre-trained on Kinetics 400, which may explain the small performance gap between our model and slowfast with non-local \cite{feichtenhofer2018slowfast}. 
\subsubsection{Something-Something V2}
Something-something dataset is unique in that it shows people manipulating objects, rather than performing common actions. We believe that distinguishing those manipulation actions needs more motion context than other datasets so it's a good dataset to validate the effectiveness of temporal modelling.
Following previous works \cite{lin2018temporal}. during inference, we use the center crop of size $224 \times 224$ from 8 segments to compute the classification score on the validation set. 

\begin{table}[t!]
\caption{Results on Something-something V2 validation set. The results are generated by taking the center crop of 1 clip/video \cite{lin2018temporal} as input to the network.}
	\begin{center}
		\begin{tabularx}{1\columnwidth}{l|c|c|c} 
		    \toprule
			Model        & Init. & FLOPs & Top1  \\ 
			\midrule
			I3D50   & K400   & 33G & 50.0  \\
			TRN \cite{zhou2018temporal}   & ImgNet   & 42G & 55.5  \\
			CIDC \cite{li2020directional}   & K400   & 92G & 56.1  \\
			TSM \cite{lin2018temporal}   & K400   & 33G & 59.1  \\
			SF \cite{wu2020multigrid}  & K400   & 66G & 60.9  \\
			SF (multigrid) \cite{wu2020multigrid}  & K400  & 66G & 61.2  \\
			TPN \cite{yang2020temporal}   & K400   & 56G & 62.0  \\
		    \midrule
			NUTA network I3D-50                 & K400 & 49G & 61.5 \\
			NUTA network I3D-101                & K400 & 98G & 62.1 \\
			NUTA network I3D-101                & K700 & 98G & 63.0 \\
			\bottomrule
		\end{tabularx}
	\end{center}
	\label{tab:ss_res}
\end{table}

As the results in Table \ref{tab:ss_res} show, our model achieves state-of-the-art recognition performance (single model with RGB input) on something-something v2 dataset with 63.0\% top1 accuracy.
Our NUTA network is able to outperform I3D and I3D based approaches \cite{wu2020multigrid, li2020directional} with lower computational cost.
It is worth mentioning that our NUTA network generalizes well to different datasets.
Previous works have demonstrated that I3D is not as effective in modeling the temporal relations \cite{lin2018temporal, li2020directional}.
The state-of-the-art methods on Kinetics e.g. TPN \cite{yang2020temporal} and slowfast \cite{yang2020temporal} must either use the TSM \cite{lin2018temporal} backbone or the multigrid training trick to achieve high performance on the Something-something dataset, showing that for Something-something dataset a specialize adaptation is usually required to transfer a high performing Kinetics model over to the task.
We apply our NUTA method directly to the unmodified I3D backbone and are able to get a 11.5\% performance boost over the I3D baseline. 
We believe this results demonstrate that our selection of informative features is not only useful in improving performance, it generalizes more easily to different domains of video understanding.

\subsubsection{Charades}
Charades is a dataset \cite{sigurdsson2016hollywood} of longer duration video annotated for a multi-label action classification problem. 
Table \ref{tab:chad_res} shows our model achieves very competitive recognition accuracy that is comparable to state-of-the-art methods while at significantly lower computational cost.
Compared to both slowfast-101 and slowfast-101 with non-local \cite{feichtenhofer2018slowfast}, we achieve slightly lower performance (41.2 mAP) but with 18\% fewer FLOPs. 
We achieve state-of-the-art performance when we initialize the model with K700 pre-trained weights, but still at much lower computational cost comparing with previous methods \cite{wu2019long, feichtenhofer2018slowfast}.

\begin{table}[t!]
\caption{Results on Charades dataset. The results are generated from an input of 32 frames sampled 1 out of 4 frames.}
	\begin{center}
		\begin{tabularx}{0.9\columnwidth}{l|c|c|c} 
		    \toprule
			Model        & Init. & FLOPs & mAP  \\ 
			\midrule
			Non-local \cite{wang2018non}   & K400   & 544G & 37.5  \\
			STRG \cite{wang2018videos}   & K400  & N/A & 39.7  \\
			LFB-NL \cite{wu2019long}  & K400   & 529G & 42.5  \\
			SF \cite{feichtenhofer2018slowfast}  & K400   & 213G & 42.1  \\
			SF-NL \cite{feichtenhofer2018slowfast}  & K400   & 234G & 42.5  \\
			X3D-XL  \cite{feichtenhofer2020x3d} & K400  & 48.4G & 43.4  \\
		    \midrule
			NUTA network 101                & K400 & 186G & 41.2\\
			NUTA network 101                & K700 & 186G & 43.1 \\
			\bottomrule
		\end{tabularx}
	\end{center}
	\label{tab:chad_res}
\end{table}

\begin{table*}[h!]
\caption{ Ablation studies on Kinetics 400 dataset. 
    We use an I3D-50 backbone. I3D and I3D* stand for the I3D network without and with temporal down sampling respectively.
	The evaluation is performed on 30 views with 32 frame input unless specified.}
    \footnotesize
	\centering
	\subfloat[Comparison between I3D with and without temporal down-sampling, and our NUTA network]{
		\begin{tabularx}{0.35\textwidth}{c|c|c} 
			\toprule
			Model        & GFLOPs & Top1   \\ 
			\midrule
			I3D* \cite{wang2018non}  &  33  & 74.0    \\ 
			I3D  \cite{yang2020temporal}     &  168  & 75.7  \\ 
			NUTA network I3D & 197 & 77.2 \\
			\bottomrule
		\end{tabularx}
	} \hfill
	\subfloat[NUTA network performance with different feature extraction strategies]{
		\begin{tabularx}{0.31\textwidth}{c|c} 
			\toprule
			Model         & Top1   \\ 
			\midrule
			I3D (backbone)  &   75.7    \\ 
			+ NUTA (only I3D features)    & 76.3  \\ 
			+ I3D and NUTA features  & 77.2 \\
			\bottomrule
		\end{tabularx}
    } \hfill
    \subfloat[Model performance with different input clip lengths]{
	\begin{tabularx}{0.22\textwidth}{c|c|c} 
		\toprule
		Input        & GFLOPs & Top1   \\ 
		\midrule
		$8\times8$  &  49  & 75.5    \\ 
		$16\times4$  & 98 & 76.8 \\
		$32\times2$  & 197 & 77.2 \\
		\bottomrule
	\end{tabularx}
    } \hfill
    \subfloat[Model performance with different input frame resolution ]{
	\begin{tabularx}{0.25\textwidth}{c|c|c} 
		\toprule
		Input        & GFLOPs & Top1   \\ 
		\midrule
		$128^2$  &  52  & 76.3    \\ 
		$256^2$  &  197  & 77.2    \\ 
		$312^2$  & 395 & 77.6 \\
		\bottomrule
	\end{tabularx}
    } \hfill
    \subfloat[Model performance with different NUTA group head configurations]{
	\begin{tabularx}{0.31\textwidth}{c|c|c} 
		\toprule
		Backbone & Number of groups        & Top1   \\ 
		\midrule
		I3D* & 16  &   75.6    \\ 
		I3D* & 32     &   76.0  \\ 
		I3D* & 64 &  76.2 \\
		\bottomrule
	\end{tabularx}
    } \hfill
    \subfloat[Performance comparison of including different NUTA stages]{
	\begin{tabularx}{0.36\textwidth}{c|c|c} 
		\toprule
		Backbone & Add NTUA at         & Top1   \\ 
		\midrule
		I3D & NUTA3+NUTA4+NUTA5   & 76.7    \\ 
		I3D & NUTA4+NUTA5    & 77.2  \\ 
		I3D & NUTA5      & 75.7 \\
		\bottomrule
	\end{tabularx}
    } \hfill
	\label{tab:ablation}
\end{table*}

\subsection{Ablation Study}
\label{sec:abla}
\subsubsection{NUTA vs. uniform temporal aggregation} We compare the performance of an I3D model with temporal down-sampling (temporal pooling and striding ) \cite{wang2018non}, an I3D without temporal down-sampling \cite{yang2020temporal} and an I3D with our proposed NUTA network (Table \ref{tab:ablation}(a)).
The results show that the I3D without temporal down-sampling outperforms the I3D with temporal downsampling. We believe this is because it is able to maintain more temporal information. 
Our NUTA network is able to further outperform I3D without temporal down-sampling by an additional 1.5\%, which shows that non-uniformly aggregating the informative temporal information helps with the action understanding.
Note that the NUTA models only add roughly 10\% additional GLOPs since the NUTA modules perform their own temporal down-sampling, significantly reducing its computational overhead. 

\subsubsection{NUTA network branch analysis} 
To understand the feature from both the uniform and non-uniform branches we look at different configurations of our network using the I3D backbone as a starting point.
First we take the baseline by using the I3D backbone as the uniform branch without adopting an non-uniform branch. 
Then we construct a NUTA network by attaching our NUTA modules to form a non-uniform branch but still perform classification from only the features coming out of the uniform branch (I3D features). 
Such setup outperforms the baseline by 0.7\% (Table \ref{tab:ablation}(b)), and we conjecture this is because the features from uniform branch utilize the temporal mapping learned by the non-uniform branch from NUTA module to drop non-informative frames. Adding the features from non-uniform branch  classification further improves the Top1 accuracy by 0.9\%, showing that those two branches learn features that are complementary for better action recognition.

\subsubsection{Frame rate} We test the NUTA network at different frame-rates following previous papers \cite{feichtenhofer2018slowfast, wang2018non, yang2020temporal}. Our results (Table \ref{tab:ablation}(c)) shows that dense sampling gives a better performance but requires significantly higher computational cost. Our method particularly benefits from more densely-sampled frames as this prevents information loss at the early stages of the network, thus giving our NUTA units sufficient information for temporal feature aggregation.  

\subsubsection{Frame resolution} Table \ref{tab:ablation}(d) compares our NUTA network performance across different input resolutions. Generally a larger input resolution leads to a better performance, but will significantly increase the computational cost. We used $256^2$ for all other experiments and GFLOPs estimation.

\subsubsection{NUTA configuration} We study the impact of number of groups used in the 3D convolutions (equation \ref{eq:temp_conv}) in our NUTA module. Our results (Table \ref{tab:ablation}(e)) show that 64 groups give the best performance. 

\subsubsection{Number of NUTA units} Table \ref{tab:ablation}(f) compares including (or not including) different NUTA stages. 
The results show that adding more NUTA stages does not neccessarily lead to better performance. Adding NUTA stage 4 and 5 gives the best performance (Table \ref{tab:ablation}(f)).
A possible explanation is that the NUTA module aggregates features by focusing on important temporal instances but stacking too many NUTA modules will compress feature too aggressively (since each NUTA will down-sample temporal dimension by 2). 

\section{Visualization}
\label{sec:vis}
\begin{figure}[t]
\centering
\includegraphics[width=0.95\columnwidth]{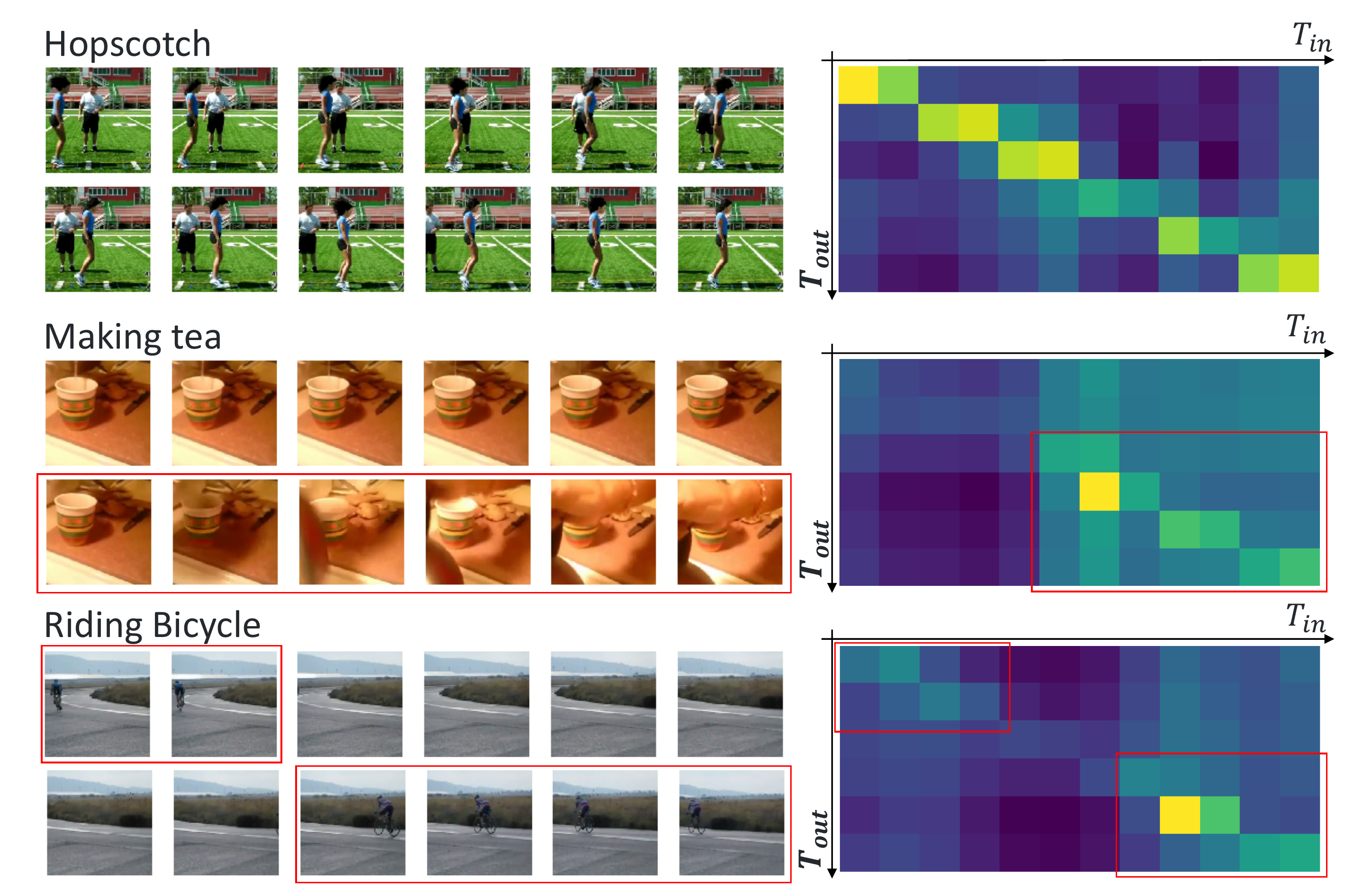}
\caption{Visualization of the temporal projection map learned by our NUTA module. Our NUTA module performs uniform aggregation when information is smoothly distributed across clip (top clip) but is able to focus on informative temporal segments when there is irrelevant segments in the video (middle and bottom clips).}
\label{fig:vis}
\end{figure}
To understand how the proposed NUTA unit works to aggregate temporal features, we visualize the temporal projection matrix (equation \ref{eq:temp_conv}).
We notice that: when the input clip has smooth transition, which indicates the information is roughly uniformly distributed over time, the NUTA unit performs uniform sampling (Figure \ref{fig:vis} top row).
Most previous works based on 3D convolutions also perform temporal information aggregation in this way.
When the frames are highly repetitive or contain only background without useful information for action recognition, the NUTA unit is able to skip the non-informative frames and focus more on the representative features (e.g. Figure \ref{fig:vis} middle row, information from first half of the clip is skipped).
When the input features have scene changes or contain noises (e.g. transition frames), the temporal mapping generated by NUTA stays to focus on representative information by skipping noises  (e.g. Figure \ref{fig:vis} bottom row, the NUTA is able to focus on the start and end of the video while skipping the middle frames). 

\section{Conclusion}
\label{sec:conclusion}
In this paper, we have addressed the ``where to extract feature" question by proposing a non-uniform temporal aggregation (NUTA) module, which is able to select the informative features.
We have further proposed the two-branch network with a uniform branch that learns local feature and non-uniform branch that learns clip-level features.
Our experimental results have showed that the proposed NUTA network achieves state-of-the-art accuracy on four public action recognition datasets. Besides, we have demonstrated how our NUTA network works by visualizing the intermediate temporal projection matrices. One future direction is to apply the proposed NUTA to other tasks (e.g. action detection). 

{\small
\bibliographystyle{ieee}
\bibliography{cvpr21}
}

\end{document}